\documentclass[10pt, conference, compsocconf]{IEEEtran}

\usepackage{amsmath,amssymb,amsfonts}
\usepackage{amsthm}
\usepackage{algorithm,algorithmicx, algpseudocode}
\usepackage{threeparttable}
\usepackage{graphicx}
\usepackage{textcomp}
\usepackage{xcolor}
\usepackage{subfigure}
\usepackage{booktabs} %
\usepackage[numbers,sort&compress]{natbib}
\usepackage{hyperref}
\usepackage{cleveref}
\usepackage{enumerate} 

\newtheorem{lemma}{Lemma}
\newtheorem{theorem}{Theorem}
\makeatletter
\newcommand{\defeq}{\mathrel{\aban@defeq}}
\newcommand{\aban@defeq}{%
  \vbox{\offinterlineskip\check@mathfonts
    \ialign{\hfil##\hfil\cr
      \fontsize{\ssf@size}{\z@}\normalfont def\cr
      \noalign{\kern1\p@}
      $\m@th=$\cr
      \noalign{\kern-.5\fontdimen22\textfont2}
    }%
  }%
}
\makeatother
\begin{document}
%
\title{WD3: Taming the Estimation Bias in Deep Reinforcement Learning\\
	\thanks{Correspondence to: Xinwen Hou xwhou@nlpr.ia.ac.cn} 
}

\author{
	\IEEEauthorblockN{Qiang He\IEEEauthorrefmark{1}\IEEEauthorrefmark{2}, Xinwen Hou\IEEEauthorrefmark{1}}
	
	\IEEEauthorblockA{\IEEEauthorrefmark{1}Institute of Automation, Chinese Academy of Sciences, Beijing, China} 
	\IEEEauthorblockA{\IEEEauthorrefmark{2}School of Artificial Intelligence, University of Chinese Academy of Sciences, Beijing, China} 
	qianghe97@gmail.com, xwhou@nlpr.ia.ac.cn
}

\maketitle

\begin{abstract}
The overestimation phenomenon caused by function approximation is a well-known issue in value-based reinforcement learning algorithms such as deep Q-networks and DDPG, which could lead to suboptimal policies. To address this issue, TD3 takes the minimum value between a pair of critics. In this paper, we show that the TD3 algorithm introduces underestimation bias in mild assumptions. To obtain a more precise estimation for value function, we unify these two opposites and propose a novel algorithm \underline{W}eighted \underline{D}elayed \underline{D}eep \underline{D}eterministic Policy Gradient (WD3), which can eliminate the estimation bias and further improve the performance by weighting a pair of critics. To demonstrate the effectiveness of WD3, we compare the learning process of value function between DDPG, TD3, and WD3. The results verify that our algorithm does eliminate the estimation error of value functions. Furthermore, we evaluate our algorithm on the continuous control tasks. We observe that in each test task, the performance of WD3 consistently outperforms, or at the very least matches, that of the state-of-the-art algorithms\footnote{Our code is available at~\href{https://sites.google.com/view/ictai20-wd3/}{https://sites.google.com/view/ictai20-wd3/}.}. 
\end{abstract}


\IEEEpeerreviewmaketitle

\section{Introduction}

The goal of reinforcement learning (RL) is to learn good policies for sequential decision-making problems by optimizing the cumulative reward signals. Combined with deep learning (DL), a lot of achievements have been produced in a wide range of fields such as playing Atari games \citep{nature_dqn}, playing chess, Go and shoji \cite{alpha}, beating human players in StarCraft \cite{Starcraft}, controlling robotic manipulation \cite{SAC}, etc. However, there still exist several severe issues that prevent deep reinforcement learning (DRL) from being applied to a wider range of tasks. One of the trickiest issues is the systematic estimation bias of value function in value-based reinforcement learning algorithms, such as Deep Q-networks \cite{Q-learning}, DDPG \cite{ddpg}, and TD3 \cite{td3}.

In typical discrete continuous control tasks, the overestimation issue of the value function has been well-studied~\citep{hasselt2010double,ddqn}. The overestimation phenomenon occurs when the value estimated by a function approximation which is larger than the true value. Overestimation bias is a property of the max operator of Q-learning, where maximization of value function estimation with noise leads to consistent overestimation bias \citep{thrun1993issues}. In the function approximation setting, noise is unavoidable which may be caused by model bias, inaccurate approximation error function, data noise, etc. This estimation error is further amplified by the nature of dynamic programming, where the value function is updated by the estimates of subsequent value, which is known as the accumulation of error \cite{rl}. It is possible to have a relatively high value for any state, such as bad states or states with few visits due to the overestimation. Inconsistent estimation bias can also destroy the quality of the gradient of the policy function for actor-critic methods because the update of policies depends on the accurate estimation of value functions. Thus, estimation bias could lead to sub-optimal policies or divergence \cite{bcq}. A more accurate value estimation can further improve the DRL agent. \citet{td3} showed that overestimation problems often occur in algorithms that use only one critic. Thus they utilize a pair of critics at the same time and take the minimum of them. This method, however, results in underestimation problems \cite{wu2020reducing} as we theoretically show in the subsequent section. The study of the underestimation bias in continuous control tasks has received very little attention \cite{lan2019maxmin,wu2020reducing}. In this paper, we focus on the problem of underestimation of value functions in continuous action space.


Our insight is if we can combine these two opposites (overestimation and underestimation) then we can get a more accurate estimation for the value function. To tackle the estimation issue, we explore how to combine these two opposite biases to make value function estimation more accurate. We adopt an easy-to-follow method. Specifically, we add an offset term to the target function. Thus, we propose a novel algorithm that offers a more accurate estimation of the value function called \underline{W}eighted \underline{D}elayed \underline{D}eep \underline{D}eterministic Policy Gradient (WD3) algorithm., which offers a kind of convex joint of underestimation and overestimation and thus somehow offers a trade-off between overestimation and underestimation. We evaluate WD3 on OpenAI gym continuous control tasks \cite{gym}, and WD3 matches or outperforms all other algorithms we tested.

Our contributions are summarized as follows.
\begin{itemize}
	\item We theoretically demonstrate that underestimation bias occurs when taking the minimum of a pair of action-value functions. Furthermore, we experimentally verify that the underestimation bias does occur and hurt performance in continuous control tasks.
	\item We propose a novel deep reinforcement learning for tackling estimation bias, namely \underline{W}eighted \underline{D}elayed \underline{D}eep \underline{D}eterministic Policy Gradient (WD3) algorithm, which utilize a novel convex connection mechanism for a pair of action-value functions for computing target value function. The convergence guarantee of WD3 is given. And we experimentally show that WD3 does tame the estimation bias.
	\item To demonstrate the effectiveness of our method, we perform experiments on gym continuous control tasks. The results show that WD3 performs better than the state-of-the-art algorithms on all OpenAI gym environments tested through more accurate action-value estimation.
\end{itemize}
\section{Related Work}
\citet{thrun1993issues} proposed that there exists an overestimation bias in Q-learning. Some researchers tried to minimize the accumulation of errors through the idea of an average function \cite{averaged-dqn}, adding a penalty term or a correction term to the learning process of policy \citep{fox16taming,mnih2016asynchronous}, or using a smooth value function approximation approach \citep{nachum2018smoothed}. Afterward, Double DQN is proposed to solve the overestimation \cite{vanhasselt10double, ddqn} problem. \Citet{vanhasselt10double} noticed that the overestimation problem often happens when using a single Q-function, so they introduced a pair of Q-functions to solve the overestimation problem and they proposed the double DQN algorithm. Double DQN reduces the overestimation of the Q-function in discrete action space, which improves performance by using two decoupled functions. Unfortunately, Double DQN still overestimates the Q value for continuous control tasks \citep{td3}. Thus, \citet{td3} proposed TD3 to reduce overestimation by taking the minimum value of a pair of action-value functions. This min operator is efficient for reducing overestimation bias. However, the min operator presents an underestimation issue, i.e., the action-value output by function approximation is lower than the true value~\citep{wu2020reducing}. Although this error does not show propagation when the value function is updated, it still makes agents pessimistic about the future because of the underestimation of the action-value function, as a result, harming the performance of algorithms. ~\citet{wu2020reducing} discussed how to leverage three value functions to alleviate the underestimation. The difference in our work is that we still use only two critics to deal with estimation bias under different assumptions. \Citet{bcq,bcqbechmarking} proposed BCQ, which still utilizes two action-value functions updated by a weighted target Q-function. However, they concentrated on how to learn control from offline fixed datasets produced by DRL methods without more discussion about the underestimation phenomenon.

\section{Preliminaries}
\label{sec: preliminaries}
We formulate the standard RL framework as a Markov Decision Process (MDP). This MDP is represented by the tuple $(\mathcal{S, A, R, } p, \rho_0, \gamma)$, encompassing the state space $\mathcal{S}$, action space $\mathcal{A}$, reward function $\mathcal{R: S \times A} \rightarrow \mathbb{R}$, transition probability function $p$, initial state distribution $\rho_0$, and a discount factor $\gamma \in [0,1]$. At each timestep $t$, the agent receives a state $s \in \mathcal{S}$ and selects an action $a \in \mathcal{A}$ according to its policy $\pi : \mathcal{S} \rightarrow \mathcal{A}$, subsequently receiving a reward $r$ and transitioning to a new state $s'$ in the environment. The return is defined as the cumulative discounted reward $R_t = \sum_{i=t}^T\gamma^{i-t}r(s_i, a_i)$ with a discounted factor $\gamma$ determining the priority of short-term rewards. Note that the return depends on the actions, and thus on the policy $\pi$, deterministic or stochastic. A trajectory $\tau = (s_0, a_0, s_1, a_1, ...)$ is a sequence of states and actions where $s_0 \sim \rho_0$ and $ a_i \sim \pi$. A transition is a tuple $(s,a,r,s')$, where action $a$ is performed at state $s$, getting reward $r$ and next state $s'$. The goal of RL is to find an optimal policy that maximizes the cumulative discounted reward $R_t$. In value-based reinforcement learning algorithms, the action-value function, a.k.a. Q-function, critic, is defined as $Q(s,a) = \mathbb{E}_{\tau \sim \pi} [R(\tau) | s_o=s, a_o=a]$ which measures the quality of an action $a$ given a state $s$. State-value function, a.k.a value function, is defined as $V(s) = \mathbb{E}_{\tau\sim \pi} [R(\tau)|s_0=s]$ that measures the quality of a specific state $s$. Both the value and action-value functions can be leveraged to evaluate the policy and further guide the algorithm to optimize the policy. So accurate estimation of the value function is of vital importance.

When the transition probability function is unknown, the state-value function can be recursively estimated by Bellman equation \citep{rl}:
\begin{equation}
Q(s,a) = r + \gamma \mathbb{E}_{s',a'} [Q(s',a')],
\label{eq: Q-learning}
\end{equation}
where $s' \sim p(\cdot|s,a)$ and $a' \sim \pi(s)$.

However, when using the function approximation method to estimate the action-value function, especially when using neural networks, there is a tendency to have a large variance due to the property of generalization of the neural network. Besides, there is the problem of estimation bias, which is composed of three factors, model bias, function approximation error, and data noise. In the following, we discuss some algorithms that are related to our work, which are DQN, Double DQN, DDPG, and TD3.

\textbf{DQN} leverages a multi-layered neural network to approximate the action-value function that for a given state $s$ outputs a vector of action-value $Q(s, \cdot; \theta)$ where $\theta$ represents the parameters of the neural network. To address the instability of the combination between neural networks and Q-learning, \citet{nature_dqn} proposed two important technologies: target network and experience replay. The optimal Q-function $Q^*(s, a)$ can be learned by minimizing the following loss function w.r.t. the neural network parameters $\theta$ according to Bellman equation (\cref{eq: Q-learning}):
\begin{equation}
L(\theta) = \mathbb{E}_{(s,a,r,s') \sim \mathcal{B}} \left[(y-Q(s,a;\theta))^2\right],
\label{eq: optimization objective of DQN}
\end{equation}
where $y = r+\gamma \max_{a'\in \mathcal{A}}Q(s',a';\theta')$ is target action-value which is computed by the frozen and separated network parameters $\theta'$ which is copied from learning parameters $\theta$ for every fixed time steps to decouple correlation between the learning critic and the target critic. And $\mathcal{B}$ is the replay buffer that stores the past transitions, which also reduces the correlation of sampled transitions. Both the target network and the experience replay dramatically improve the performance of DQN. Although DQN can achieve human-level control in many real-world tasks, e.g. playing Atari games, there still are some issues in this algorithm. 

\textbf{Double DQN.} A well-known issue of DQN is the overestimation of the Q-function~\citep{thrun1993issues}. The DQN algorithm involves a max operator in the construction of its target value, which makes it more likely to select overestimated values, resulting in over-optimistic value estimates in action selection. Double DQN \cite{hasselt2010double} decouples the selection from the evaluation. Double DQN is unbiased for action-value functions approximation. Although the performance of Double DQN is better than that of DQN in discrete action space, it still suffers when applied to continuous action setting \citep{td3}.

\textbf{DDPG} \citep{ddpg} leverages a Deterministic Policy Gradient method (DPG)~\citep{dpg} to optimize the expected reward which uses a deterministic policy $\pi: \mathcal{S \rightarrow A }$ instead of typical stochastic policies in the actor-critic setting with continuous action space. DPG does not need to integrate actions, so it is a more efficient method to estimate value functions than stochastic policy. DDPG leverages a learned action-value function to update the policy. We use $\phi, \phi',\theta$, and $\theta'$ to mark the parameters of the actor, target actor, critic, and target critic, respectively. The critic is updated with
\begin{equation}
L(\theta) = \mathbb{E}_{(s,a,r,s') \sim \mathcal{B}} \left[(y-Q(s,a;\theta))^2\right],
\label{eq: DDPG_Value_function_update}
\end{equation}
where $y=r+\gamma \mathbb{E}_{s'} Q(s',\pi(s';\phi');\theta')$ is the target value, computed with the independent target networks. The updating rule of the policy network is dependent on the critic with parameters $\theta$ and updated by the chain rule of gradient propagation
\begin{equation}
\nabla_\phi J(\phi) = \mathbb{E}_{s\sim p_\pi}[\nabla_a Q(s,a;\theta)|_{a=\pi(s;\phi)}\nabla_\phi\pi(s;\phi)].
\label{eq: DPG_update_rule}
\end{equation}
After updating online learning parameters, the target network parameters, $\theta'$, $\phi'$,  are soft-updated in an exponential moving average style 
\begin{equation}
\theta' = \eta \theta + (1-\eta) \theta', \; \phi'=\eta \phi + (1-\eta)\phi',
\label{eq: soft update}
\end{equation}
where $\eta$ is a small constant, controlling the magnitude of updating. The soft update remarkably improves the stability of the learning process. DPG algorithm suffers from exploration capability. 


\textbf{TD3}~\citep{td3} is an improved version of the DDPG algorithm and is also a DPG algorithm. It takes the minimum value in a pair of critics as the target value in TD3, which is called clipped double Q-learning. TD3 leverages the same form of value function loss \cref{eq: DDPG_Value_function_update}, but the target is 
\begin{equation}
 y=r+\gamma \min_{i=1,2}Q_i'(s',\pi(s';\phi');\theta'_i),  
\end{equation} 
where $Q_1'$ and $Q_2'$ represent the two target critics with respect to two independent critics $Q_1$ and $Q_2$, which alleviates the overestimation. Besides, the TD3 algorithm decouples the critic from the actor by keeping different updating frequencies for them.

\section{Diagnosing Underestimation Phenomenon}
In this section, we begin with a theoretical analysis of the underestimation bias that the occurrence of the min operator in the learning of action-value functions leads to. Then we empirically show that using the minimum value of two critics can cause underestimation bias and thus harm performance in the recently proposed TD3 algorithm. 
According to the Bellman equation (\cref{eq: Q-learning}), the learning process of the action-value function involving the min operator can be expressed as
\begin{equation}
Q(s,a) \leftarrow r+\gamma \min_{i=1,2}Q_i'(s',\pi(s';\phi');\theta'_i).
\end{equation}
To better understand the learning process, we assume that $\hat{Q_i}$ is an estimate of the true action-value $Q^*$, with an estimated error of $Z_i=\hat{Q_i}-Q^*$ due to noise, where $Z_i$ is sampled from a specific independent identical distribution. The minimization operator acts on $\hat{Q_i}$. By assuming that $Z_i$ satisfies different assumptions, we explain theoretically that the minimization operator can cause the underestimation problem of value functions.

\begin{theorem}
\label{theorem1}
Let $Q^*$ denotes the true state-action value, suppose that there are 2 estimate value $\hat{Q}_i$ for $i=1,2$ Denote the estimated error $G_i=\hat{Q}(s,a)-Q^*(s,a)$ are independently Gaussian distribution $\mathcal{N}(0,\sigma^2)$ Then,\\
	$$\mathbb{E}[\min_{i=1,2}\{G_i\}]=-\sigma\frac{1}{\sqrt{\pi}}.$$
 \end{theorem}
 
\begin{proof}
	Obviously, we have $\min\{G_1,G_2\}=\frac{1}{2}(G_1+G_2-|G1-G2|)$. Denote $Y = G_1-G_2$, hence, $Y\sim\mathcal{N}(0,2\sigma^2)$,
	\begin{equation*}
	\begin{aligned}
	\mathbb{E}[|Y|] &= \int_{-\infty}^{+\infty}|y|\phi(y)dy \\
	&= \sigma \frac{2}{\sqrt{\pi}},
	\end{aligned}
	\end{equation*}
	where $\phi(y)=\frac{1}{\sqrt{2\pi}\sqrt{2}\sigma}\exp\{-\frac{y^2}{2\cdot2\sigma^2}\}$. \\
	This implies that $\mathbb{E}[\min\{G_1,G_2\}]=\mathbb{E}[\frac{1}{2}(G_1+G_2-|G1-G2|) =-\sigma\frac{1}{\sqrt{\pi}}$.
\end{proof}
The expectation is $-\sigma \frac{1}{\sqrt{\pi}}<0$. \Cref{theorem1} reveals a surprising fact: even though the function approximation is unbiased, there still exists an \textbf{underestimation} issue. 

\begin{theorem}
\label{theorem 2}
Consider a state $s$ where true optimal action values is $Q^*$, suppose that there are $N$ estimate value $\hat{Q}_i$ for $i=1,\cdots, N.$ Denote the estimate error $Z_i=\hat{Q_i}(s,a)-Q^*(s,a)$ are independently distribution uniformly in interval $[-\delta, \delta]$. Then,\\
	$$\mathbb{E}[\min_{i=1,\cdots,N}\{Z_i\}]=-\frac{N-1}{N+1}\delta.$$
 \end{theorem}
\begin{proof}
	We denote the probability density function of $Z_i$ for $i=1,\cdots,N$ as $f(x)$:
	\begin{equation*}
	\begin{aligned}
	f(x) = \left\{ \begin{array}{rcl}
	\frac{1}{2\delta},	&	& {x \in (-\delta, \delta)} \\
	0,    &    & {else.}\\
	\end{array}\right.
	\end{aligned}
	\end{equation*}
	Then, we can derive the cumulative distribution function for all variables $Z_i$, where $i=1,\cdots, N$:
	\begin{equation*}
	\begin{aligned}
	P\{Z_i>x\}&= \left\{ \begin{array}{rcl}
	1,	&	& {x \leq -\delta} \\
	\frac{\delta - x}{2\delta},    &    & {x \in (-\delta, \delta)}\\
	0,	&	& { x \geq \delta.} \\
	\end{array}\right.
	\end{aligned}
	\end{equation*}
	Since $Z_i$ is a uniformly random variable in $[-\delta, \delta]$, the probability that $\min_{i=1,\cdots,N}Z_i \geq x$ for $x$ is equal to the probability that $Z_i \geq x$ for all $i=1,\cdots,N$ simultaneously, we can derive:
	\begin{equation*}
	\begin{aligned}
	P\{ \min_i Z_i\geq x \} &= P\{Z_1\geq x, Z_2\geq x, \cdots, Z_N\geq x\} \\
	&= \prod_{i=1}^{N}P\{Z_i\geq x\} \\
	&= \left\{ \begin{array}{rcl}
	1,	&	& {x \leq -\delta} \\
	(\frac{\delta - x}{2\delta})^N,    &    & {x \in (-\delta, \delta)}\\
	0,	&	& {x \geq \delta.} \\
	\end{array}\right.
	\end{aligned}
	\end{equation*}
	This implies that we can get the cumulative density function (CDF):
	\begin{equation*}
	\begin{aligned}
	P\{ \min_{i=1,\cdots,N} Z_i < x \} &= 1 - P\{  \min_{i=1,\cdots,N} Z_i\geq x \} \\
	&= \left\{ \begin{array}{rcl}
	0,	&	& {x \leq -\delta} \\
	1-(\frac{\delta - x}{2\delta})^N,    &    & {x \in (-\delta, \delta)}\\
	1,	&	& {x \geq \delta} \\
	\end{array}\right.
	\end{aligned}
	\end{equation*}
	Then, we can get the probability density function of this variable by using the derivative of the CDF:
	\begin{equation*}
	\begin{aligned}
	f_{min}(x) &= \frac{d}{dx}P\{ \min_{i=1,\cdots,N} Z_i < x \} = \frac{N}{2\delta}(\frac{\delta-x}{2\delta})^{N-1},
	\end{aligned}
	\end{equation*}
	for $x\in(-\delta, \delta)$. Its expectation can be written as an integral
	\begin{equation*}
	\begin{aligned}
	\mathbb{E}[\min_{i=1,\cdots,N} Z_i] &=\int_{-\delta}^{\delta}xf_{min}(x)dx \\
	&= \int_{-\delta}^{\delta}x \frac{N}{2\delta}(\frac{\delta-x}{2\delta})^{N-1}dx \\
	&= -\frac{N-1}{N+1}\delta.
	\end{aligned}
	\end{equation*}
\end{proof}
When $N=2$, the expectation is $-\frac{1}{3}\delta < 0$. \Cref{theorem 2} further extends the result of \cref{theorem1} that underestimation bias still exists in TD3 even if the bias satisfies a uniform distribution.

\subsection{Does the theoretical estimation bias occur in practice for state-of-the-art methods?}
\begin{figure}[htbp]
	\centering
	\subfigure[Overestimation in DDPG]{
		\includegraphics[width=1.6in]{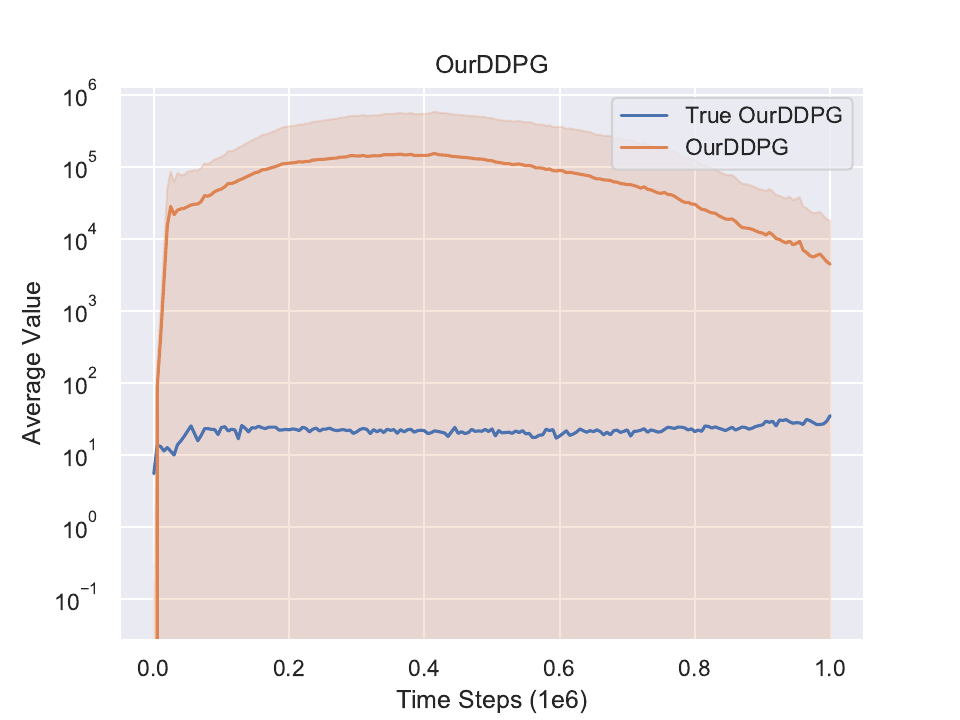}		
	}
	\hspace{-0.2in}
	\subfigure[Estimation in TD3 and WD3]{
		\includegraphics[width=1.6in]{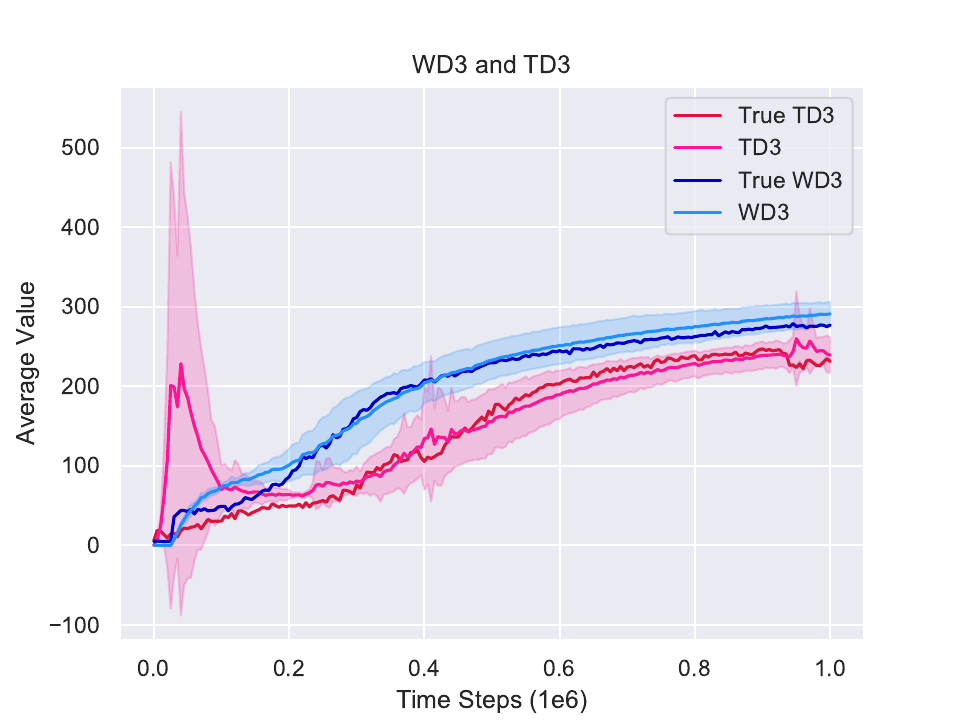}
	}
	\caption{The estimation bias on Ant-v0 PyBullet continuous control task. (a) Overestimation bias in DDPG. The estimation value of DDPG is significantly larger than that of the true value. (b) Underestimation issue in TD3 and accurate estimation in WD3. WD3 achieves a balance of these two opposites so that the estimation is more accurate than that of TD3 and DDPG. The shaded area stands for a standard deviation of the average evaluation over 10 random seeds.} 
	\label{Estimatation issue}
\end{figure}
We conduct experiments to verify whether the underestimation occurs in practice or not. We utilize Ant-v0 of the PyBullet suite \cite{bullet3} on OpenAI gym environments to verify that overestimation occurs in DDPG and TD3 does underestimate the action value. More details of the experiments are discussed in \cref{sec: implementation details}.  In \cref{Estimatation issue}, we graph the average value estimate where every data point is estimated based on 50 trajectories and compare it to the true value. At the beginning of the DDPG algorithm learning process, the Q-function is greatly overestimated, and then the Q-function slowly declines. Together with \cref{performance experiments}, we find that the overestimation of the Q-function does make the DDPG algorithm unable to learn the optimal policy. With the decrease of overestimation, the performance of DDPG increases gradually. Due to the instability of the reinforcement learning environment, the TD3 algorithm also presents the problem of greatly overestimating the initial learning process, which is then controlled by the min operator. In the later learning stage, the learning curve of the value function becomes stable, and the Q-function is underestimated and maintained for a long time. The experimental results verify the existence of overestimation and underestimation. We unify these two opposites and propose the WD3 algorithm. By weighted averaging target critic, WD3 enables the Q-function to reach a balance between overestimation and underestimation, which makes the learning process of value function estimation more stable and accurate, leading to a better policy.
\section{Weighted Delayed Deep Deterministic Policy}

To address estimation bias, Double DQN introduces a separated Q-function which still overestimates action value in continuous control tasks. TD3 takes the minimum value of two critics as a target to update the value function which results in underestimation bias. Based on TD3, we propose a novel Weighted Delayed Deep Deterministic Policy Gradient algorithm that alleviates the estimation bias by introducing a weighted smooth update mechanism that can be applied to any actor-critic algorithm.

\subsection{Weighted target update}
\label{sec: weighted target update}
In Double DQN, greedy value function update is deconstructed by keeping two Q functions, $Q_1$ and $Q_2$ can be used to update each other. However, the purpose of decoupling cannot be achieved in continuous control tasks due to the vast action space.  The slow change of the policy learning process makes the two Q networks coupled due to the slow learning process in continuous action space and the early exploration of the agent, which cannot reflect the principles that informed its development. The huge variance of Q-function estimation brought by the continuous action space makes the learning of the value function unstable compared with the discrete action state, thus leading to the overestimation problem \cite{td3}. The DDPG algorithm tends to overestimate the action-value function on the continuous control task. Fujimoto et al. proposed to use the minimum value of a pair of critics as the target for updating, resulting in an underestimation bias as discussed above.
The overestimation problem of the DDPG algorithm and the underestimation problem of TD3 are exactly two opposites. We propose the WD3 algorithm to achieve the balance between overestimating and underestimating by weighting a pair of target critics. We utilize a pair of critics, $Q_1$ and $Q_2$, and a policy network $\pi$. The parameterized Q functions are updated by:
\begin{equation}\label{eq: WD3 update rule}
\begin{aligned}
Q_i \leftarrow r + \gamma ( & \beta \min_{i=1,2}Q_i(s',a';\theta'_i) \\
&+  \frac{1-\beta}{2}\sum_{i=1}^{2}Q_i(s',a';\theta'_i) ),
\end{aligned}
\end{equation}
where $\beta \in [0,1)$ controls the balance between overestimation and underestimation. When $\beta=1$, the algorithm decays to TD3. The parameters of the actor are updated by
\begin{equation}
\hat{J(\phi)\!} =N^{-1}\!\sum_{s,a}\nabla_aQ(s,\!a;\theta_1)|_{a=\pi(s;\phi)}\!\nabla_\phi\pi(s;\phi).
\end{equation}

\begin{algorithm}
	\caption{Weighted Delayed Deep Deterministic Policy Gradient (WD3)}
	\begin{algorithmic}[1]
		\State Initialize actor network $\pi$, and critic network $Q_i$ for $i=1,2$  with random parameters $\phi, \theta_i$
		\State Initialize target networks $\theta'_i \leftarrow \theta_i$, $\phi' \leftarrow \phi$
		\State Initialize replay buffer $\mathcal{B}$
		\State Initialize $\beta$, $d$, $\sigma$, $\tilde{\sigma}$, $\eta$, $c$ total steps $T$, and $t=0$
		\State Reset the environment and receive initial state $s$
		\While{$t < T$} 
		\State Select action with noise $a = \pi(s;\phi) + \epsilon, \epsilon \sim \mathcal{N}(0, \sigma^{2}) $, and receive reward $r$, new state $s'$
		\State Store transition tuple $(s, a, r, s')$ to $\mathcal{B}$
		\State Sample mini-batch of $N$ transitions $(s, a, r, s')$ from $\mathcal{B}$
		\State $\tilde{a} \leftarrow \pi(s';\phi') + \epsilon$, $\epsilon \sim clip(\mathcal{N}(0, \tilde{\sigma}^2), -c, c)$
		\State $y \leftarrow r + \gamma (\beta\min_{i=1,2}Q(s', \tilde{a};\theta'_i)+\frac{1-\beta}{2}\sum_{i}^{2}Q(s', \tilde{a};\theta'_i))$
		\State Update critic $\theta \leftarrow N^{-1} \sum(y-Q_{\theta}(s,a))^2$
		\If{$t$ mod $d$}
		\State Update $\phi$ by the deterministic policy gradient:
		\State $\!\nabla_\phi J(\phi)\!=N^{-1}\!\sum\nabla_aQ(s,\!a;\theta_1)|_{a=\pi(s;\phi)}\!\nabla_\phi\pi(s;\phi)$
		\State Update target networks:
		\State $\theta'_i \leftarrow \eta \theta_i + (1-\eta) \theta'_i$
		\State $\phi' \leftarrow \eta \phi + (1-\eta) \phi'$
		\EndIf
		\State $t\leftarrow t+1$
		\State $s\leftarrow s'$	
		\EndWhile
	\end{algorithmic}
	\label{WD3}
\end{algorithm}
Given the Robbins-Monro stochastic approximation condition \cite{convergence}, WD3 converges to the optimal value function under the grid setting. In the subsequent section, we discuss the convergence guarantee.
\subsection{Convergence guarantee}
We are interested in the convergence property of WD3. To build the convergence result, we utilize stochastic approximation result in \citet{convergence, td3}.

\begin{lemma}[\citet{convergence}]\label{lemma: stochastic convergence}
Given a process $(\Delta_t, \alpha_t, F_t)$, $t>0$, where $\alpha_t \in [0,1]$, $\Delta_t$ and $F_t$ taking values in $\mathbb{R}^N$ and defined as $\Delta_{t+1}(x_t)=\left(1-\alpha_t(x_t)\right) \Delta_t(x_t)+\alpha_t(x_t) F_t(x_t), $
where $x_t \in \mathcal{X}$, converges to zero w.p.1 under the following assumptions:

\begin{enumerate}[(i)]
    \item The set $\mathcal{X}$ is finite.
    \item $0 \leq \alpha_t \leq 1, \sum_t \alpha_t(x)=\infty$ and $\sum_t \alpha_t^2(x)<\infty$.
    \item $\left\|\mathbb{E}\left[F_t(x) \mid \mathcal{F}_t\right]\right\|_W \leq \upsilon \left\|\Delta_t\right\|_W + \mathbf{c_t}$, with $\upsilon \in [0,1)$ and $\mathbf{c_t}$ converges to 0 with probability 1.
    \item $\operatorname{var}\left[F_t(x) \mid \mathcal{F}_t\right] \leq C\left(1+\left\|\Delta_t\right\|_W^2\right)$, for some constant $C>0$.
\end{enumerate}
\end{lemma}

The convergence guarantee of WD3 in the tabular setting is formally given in \cref{theorem: convergence of WD3}.
\begin{theorem}\label{theorem: convergence of WD3}
Given the following condition 
\begin{enumerate}[(1)]
    \item The MDP is finite. And the Q values of the two critics are stored in two tables $Q^A$ and $Q^B$. 
    \item Each state and action is sampled an infinite number of times. Both $Q^A$ and $Q^B$ are updated an infinite number of times. 
    \item $\gamma \in [0,1)$. And the learning rate $\alpha_t$ satisfy $\alpha_t \in [0, 1]$, $\sum_t \alpha_t = \infty$, and $\sum_t (\alpha_t)^2 \leq \infty$ and $\alpha_t(s,a) = 0 $ if $(s,a) \neq (s_t, a_t)$.
    \item The rewards are bounded by some constant $R_\text{max} < \infty$, i.e., $ - R_\text{max} <r(s_t,a_t) < R_\text{max}$.
\end{enumerate}
The WD3 algorithm, defined in \cref{eq: WD3 update rule}, will converge to the standard optimal value function $Q^*$ $w.p. 1$.
\end{theorem}

\begin{proof}
By the assumptions, the conditions (i) and (ii) in \cref{lemma: stochastic convergence} hold. Assumption (4) means the rewards are finite, which ensures condition (iv) holds. Next, we verify condition (iii).

Let $\mathcal{X} = \mathcal{S} \times \mathcal{A}$ and $x \in \mathcal{X}$. For a $x$, the update target of WD3 is represented as:
\begin{equation}
y = r + \gamma (\beta \min \{Q^A(x), Q^B(x)\} + \frac{1-\beta}{2} ( Q^A(x)+Q^B(x)).
\end{equation}
For notation simplicity, we omit Q table input $x$. We discuss $Q^A$ and $Q^B$ respectively. For \(Q^A\):
\begin{equation}
Q_{t+1}^A = (1-\alpha)Q_t^A + \alpha y.
\end{equation}
From this, we have:
\begin{equation}
\begin{aligned}
\Delta_{t+1}^A & \defeq Q_{t+1}^A - Q^* \\
&=  (1-\alpha)Q_t^A + \alpha y  - Q^* \\
&= (1-\alpha)(Q_t^A - Q^*) + \alpha(y - Q^*) \\ 
&= (1 - \alpha) \Delta_t^A + \alpha (y - Q^*)
\end{aligned}
\end{equation}
Let's define \(F^A\):
\begin{equation}
    \begin{aligned}
        F^A &= y - Q^* \\
            &= r + \gamma \left( \beta \min \{Q^A,Q^B\} + \frac{1-\beta}{2} (Q^A+Q^B) \right) - Q^* \\
            &\phantom{=} + \gamma Q^A - \gamma Q^A \\
            &= r + \gamma Q^A - Q^* \\
            &\phantom{=} + \gamma \left( \beta \min \{Q^A,Q^B\} + \frac{1-\beta}{2} (Q^A+Q^B) \right) - \gamma Q^A  \\
            &= F^{Q^A} \\
            &\phantom{=} +\gamma \left( \beta \min \{Q^A,Q^B\} + \frac{1-\beta}{2} (Q^A+Q^B) \right) - \gamma Q^A,
    \end{aligned}
\end{equation}
where $F^{Q^A}=r + \gamma  Q^A - Q^*$. By condition (iii), We want to ensure:
\begin{equation}
\gamma ( \beta \min \{Q^A,Q^B\} + \frac{1-\beta}{2} (Q^A+Q^B)) - \gamma Q^A \rightarrow 0, \quad t \rightarrow \infty.
\end{equation}
To see this, we further discuss the result of $\min \{Q^A, Q^B\}$.
For the case \( \min \{Q^A, Q^B\} = Q^A \):
\begin{equation}
    \begin{aligned}
& (\beta \min \{Q^A,Q^B\} + \frac{1-\beta}{2} (Q^A+Q^B)) -  Q_1 \\
= &  \beta Q^A + \frac{1-\beta}{2}(Q^A + Q^B) - Q^A \\
=& (\beta -1 ) Q^A + \frac{1-\beta}{2}(Q^A + Q^B) \\
= & \frac{\beta-1}{2}(Q^A - Q^B)
    \end{aligned}
\end{equation}
We need to ensure $Q^A - Q^B$ converges to $0$. Thus, we make a new stochastic process: $\Delta^{AB}_{t+1} = (1-\alpha_t(x_t))(\Delta_t^{AB}(x_t) + \alpha_t(x_t)F^{AB}_t(x_t)$ where $\Delta^{AB}_t = Q^A_t - Q^B _t$, where $F^{AB}$ is unknown. To find $F^{AB}$, we have
\begin{equation}
\begin{aligned}
\Delta_{t+1}^{AB} &= Q^{A}_{t+1} - Q^B_{t+1} \\
& = (1-\alpha)Q_t^A + \alpha y_t - (1-\alpha)Q_t^B - \alpha y_t \\
&= (1-\alpha)(Q_t^A -Q_t^B) \\
&= (1-\alpha) \Delta_t^{AB} + \alpha F^{AB}.
\end{aligned}
\end{equation}
Thus, $F^{AB}=0$. According to \cref{lemma: stochastic convergence},  \(\Delta^{AB} \) converges to 0. For the case \( \min Q_i = Q^B \), the same result holds, which ensures $Q^A$ converges to the optimal value $Q^*$. The convergence of $Q^B$ also holds by a similar process. Thus, condition (iii) in \cref{lemma: stochastic convergence} holds, which completes the proof.
\end{proof}
\subsection{Exploration}
A well-known issue of deterministic policy gradient algorithms is a lack of exploration capability because they directly output a certain action rather than a distribution of the action. To tackle this issue, the original DDPG proposed to increase the exploration ability by adding noise to the action which is drawn from the Ornstein-Uhlenbeck process~\citep{uhlenbeck1930theory}. However, \citet{td3} found that this kind of noise has no additional benefit to the exploration, and the same performance can be achieved with Gaussian noise. Matthias et al. added noise to the parameters of the neural network, but the method has no significant advantage over the former one in the continuous control tasks~\cite{plappertparameter}. To ensure exploration, we add Gaussian noise to actions when the agent interacts with the environment and target actions. Therefore, our target action $\hat{a}$ is:
\begin{equation}
\hat{a} = a + \epsilon, 
\end{equation}
where $a = \pi(s;\phi) $, $\epsilon \sim \mathcal{N}(0, \sigma^2)$. We summarize the algorithm in \cref{WD3}.

\begin{table*}[htbp]
	\caption{The average of the last 5 returns over 10 trials of 1 million time steps for various $beta$. The maximum value for each task is bolded. + corresponds to a single standard deviation over trials.}
	\renewcommand\tabcolsep{3.0pt} 
	\begin{center}
		\begin{tabular}{lllllllll}
			\toprule
			\textbf{$\beta$}&\textbf{Ant}&\textbf{HalfCheetah}&\textbf{Hopper}&\textbf{InvertedDouble}&\textbf{InvertedPendulum}&\textbf{Swingup}&\textbf{Reacher}&\textbf{Walker2D} \\
			\midrule
			0.15&2810.89+183.66&2616.87+300.32&2113.79+292.63&9321.41+153.8&985.84+70.8&887.93+13.97&\textbf{20.14+2.39}&1760.08+298.9\\
			0.30&2948.46+178.79&2551.4+253.1&2101.9+248.49&8691.23+1718.2&\textbf{1000.0+0.0}&870.81+42.02&19.53+2.87&1773.36+230.79\\
			0.45&2868.36+257.58&\textbf{2630.83+139.69}&2115.72+282.72&9023.9+1223.6&845.31+294.0&886.67+13.63&19.03+2.64&\textbf{1943.26+190.86}\\
			0.50&2964.19+128.02&2261.6+176.84&\textbf{2215.27+200.02}&8775.14+1346.84&969.62+151.92&\textbf{889.3+0.83}&19.08+3.24&1798.14+263.92\\
			0.60&2813.11+206.33&2476.65+204.58&1768.56+742.98&9022.89+1244.94&942.89+204.14&885.13+19.08&18.45+4.26&1841.39+227.03\\
			0.75&\textbf{3042.37+100.38}&2285.7+296.99&1978.8+189.18&\textbf{9323.49+166.4}&\textbf{1000.0+0.0}&886.2+11.89&19.65+3.36&1625.85+407.89\\
			TD3&2741.59+277.94&2358.92+227.99&1850.97+590.81&6544.22+4239.66&773.24+411.3&887.64+8.47&18.71+3.17&1674.36+340.4\\
			\bottomrule 
		\end{tabular}
		\label{various_beta}
	\end{center}
\end{table*}
\section{Experiments}

\begin{figure*}[!ht]
	\centering
	\subfigure{
		\includegraphics[width=1.8in]{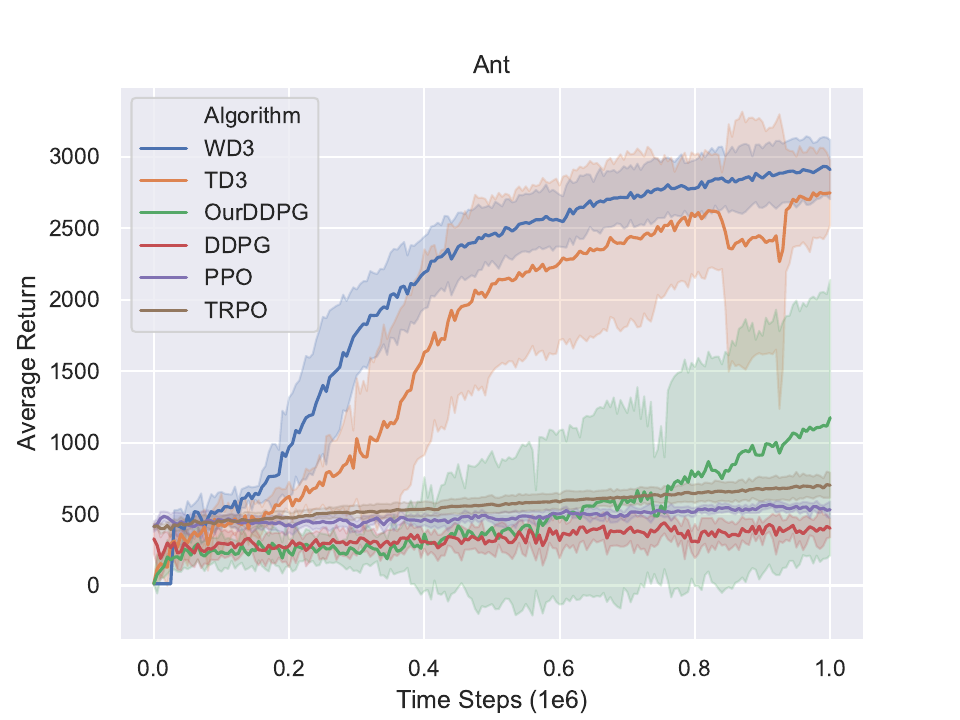}
	}
	\hspace{-0.3in}
	\subfigure{
		\includegraphics[width=1.8in]{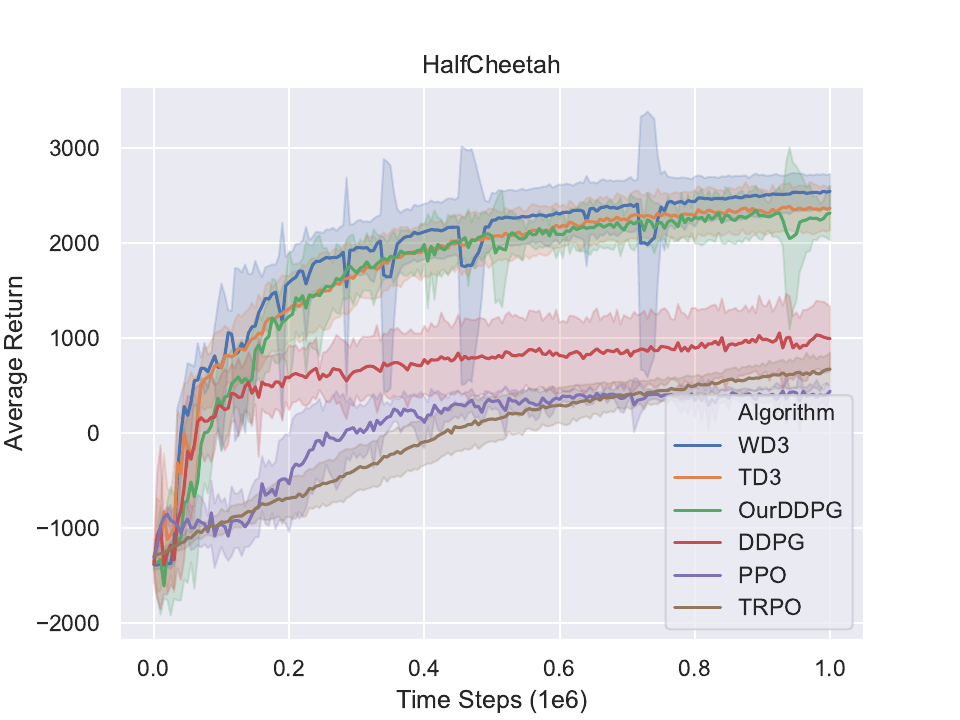}
	}   
	\hspace{-0.3in}
	\subfigure{
		\includegraphics[width=1.8in]{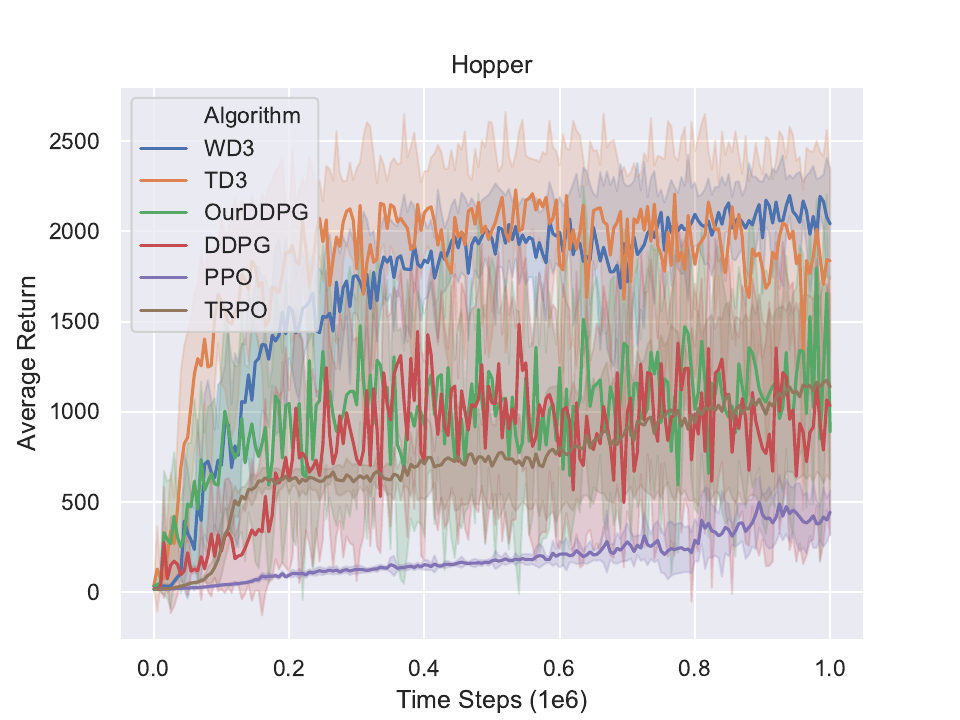}
	}
	\hspace{-0.3in}
	\subfigure{
		\includegraphics[width=1.8in]{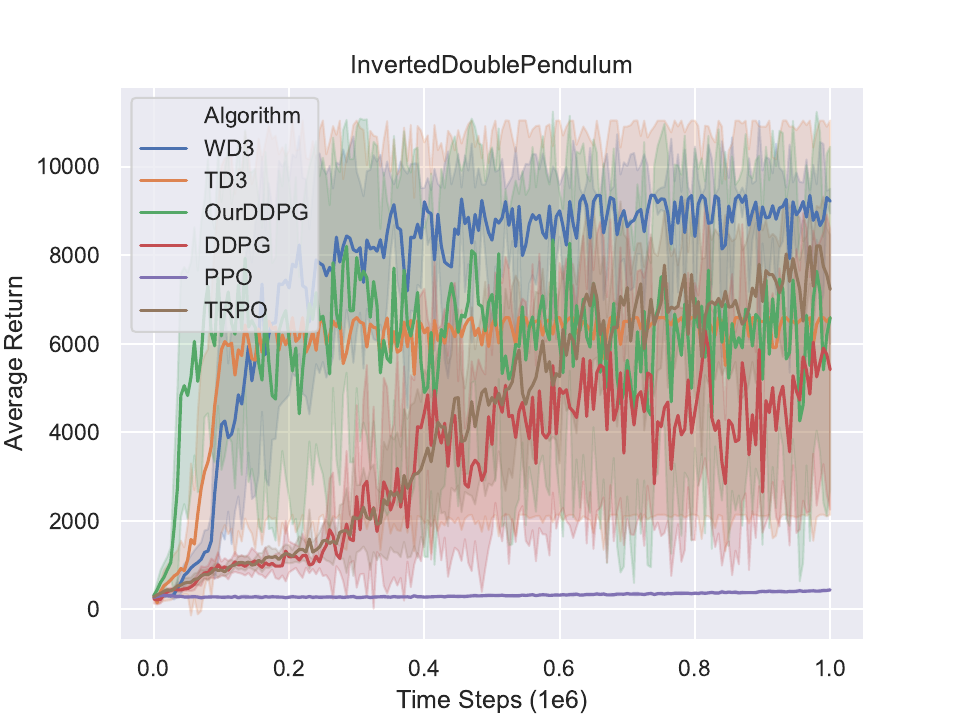}
	}
	\hspace{-0.3in}
	\subfigure{
		\includegraphics[width=1.8in]{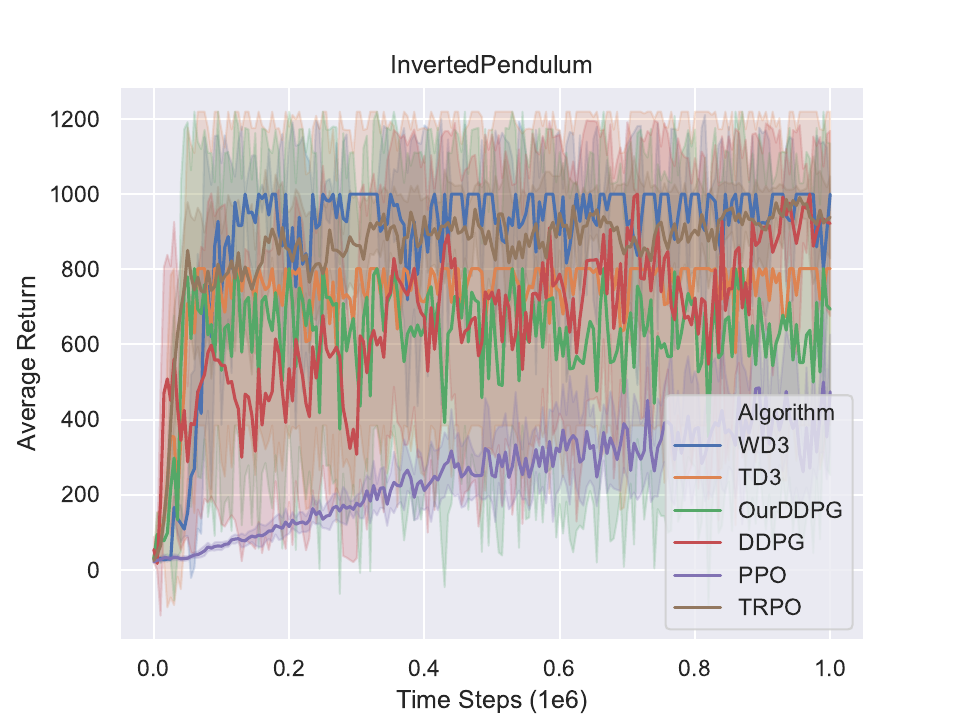}
	}
	\hspace{-0.3in}
	\subfigure{
		\includegraphics[width=1.8in]{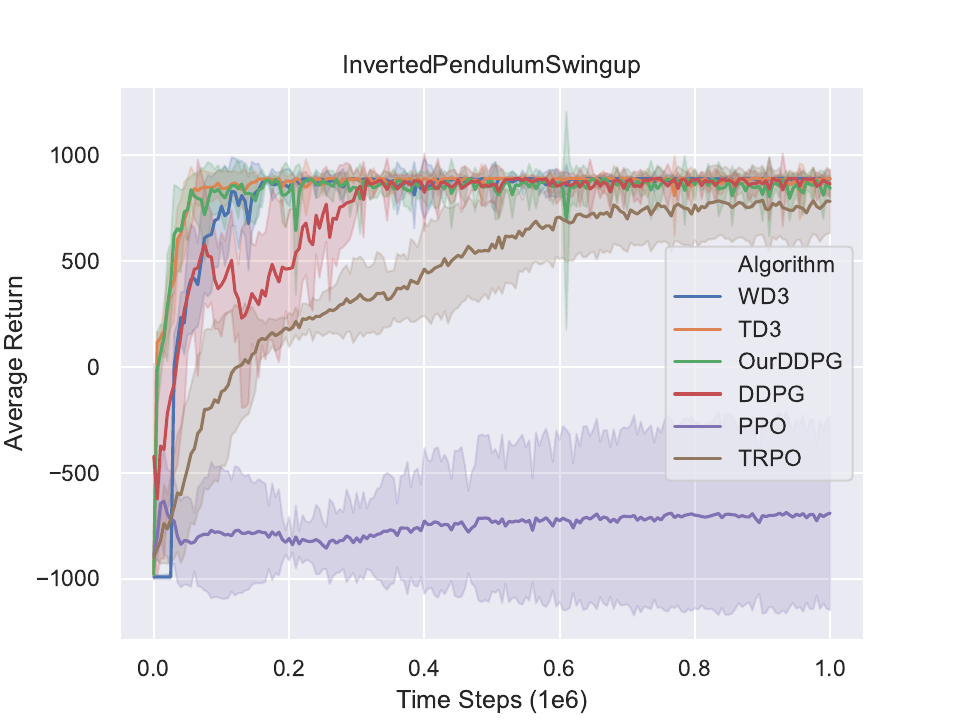}
	}   
	\hspace{-0.3in}
	\subfigure{
		\includegraphics[width=1.8in]{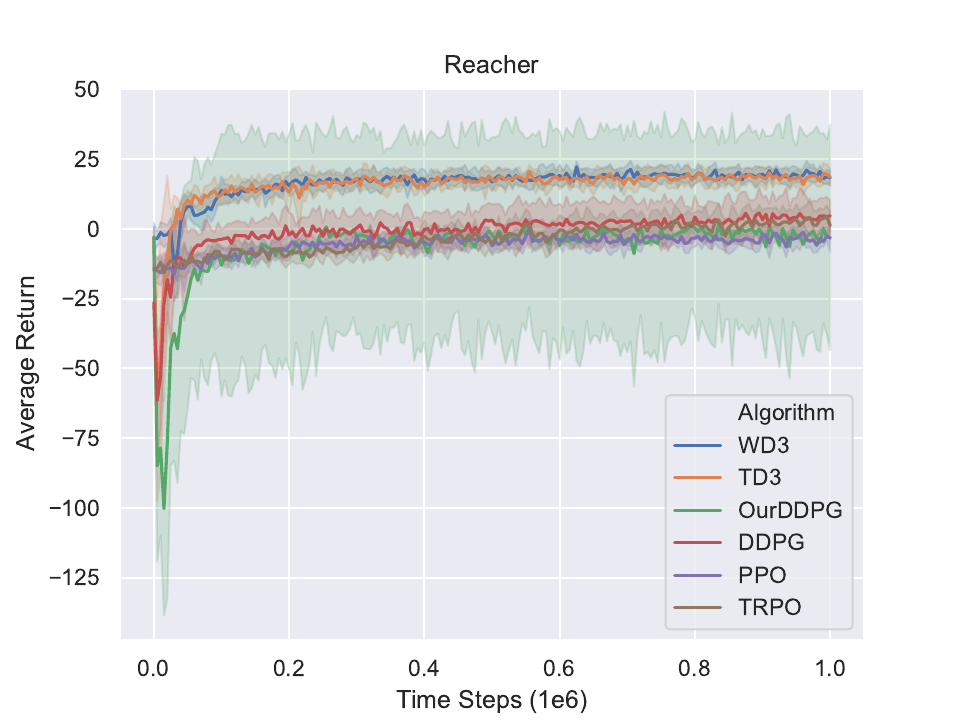}
	}
	\hspace{-0.3in}
	\subfigure{
		\includegraphics[width=1.8in]{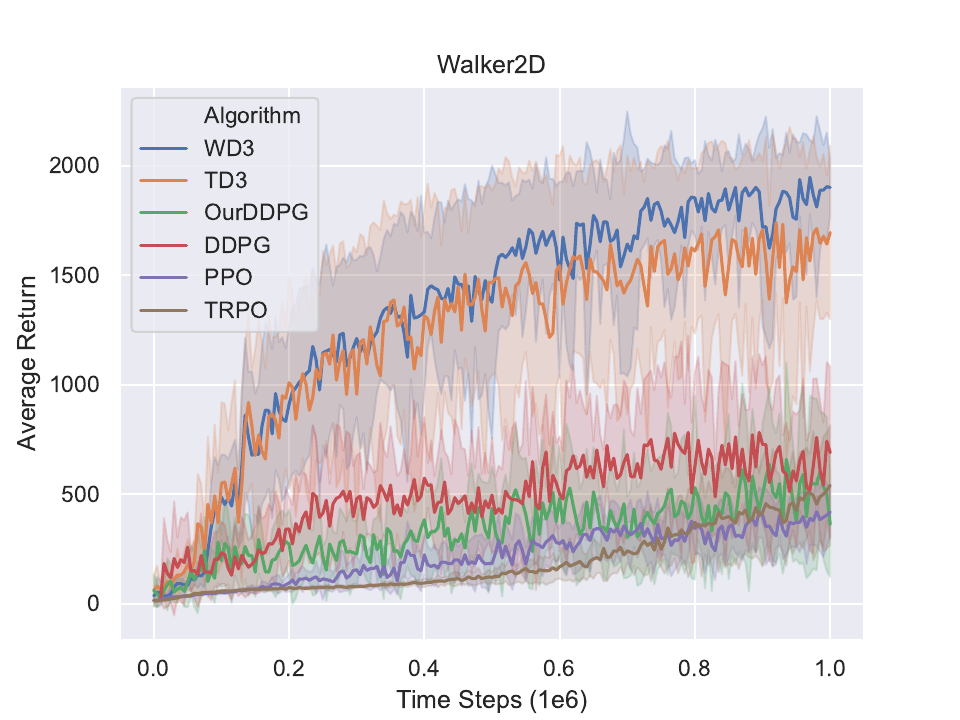}
	}
	\caption{Performance curves for OpenAI gym continuous control tasks in PyBullet suite. WD3 outperforms the other tested algorithms. The shaded region represents a standard deviation of the average evaluation over 10 seeds. The curves are not smooth at all.}
	\label{performance experiments}
\end{figure*}
To evaluate our algorithm, we measure the performance of WD3 on the suite of PyBullet \cite{bullet3} continuous control tasks. By using the modifications discussed in \cref{sec: weighted target update}, we increase the stability and accuracy of the Q-function learned by considering the estimation problem. The WD3 algorithm still maintains a pair of critics, and we update the Q-function with a weighted average \cref{eq: WD3 update rule}.
The policy network is updated by \cref{eq: DPG_update_rule}. Utilizing a specific action-value function, we can update the parameters of the policy network through the chain rule of gradient propagation. Every $d$ time step, the policy network is updated with the action-value function, according to the deterministic policy gradient algorithm \citep{dpg}. To increase the stability and performance of the algorithm, the soft update method is adopted when updating target networks.

\subsection{\label{sec: implementation details}Implementation Details}
Given the recent concerns about algorithms reflecting the principles that informed its development \cite{drl_matters}, we implement WD3 simply without any engineering tricks to make the algorithm work as we originally intended. We use the original low-dimensional state vector provided by the environment as input without any modification.  Besides, we use the default reward functions and environment settings without any changes to achieve a fair comparison of performance.

Due to recent concerns about the reproducibility crisis of deep reinforcement learning algorithms \cite{drl_matters}, we run all the tested algorithms over 10 random seeds. For WD3, we use a two-layer feedforward neural network, each layer has 256 units, using the rectified linear units (ReLU) as the activation function of each layer for all actors and critics, but the last layer of the actor is followed by a tanh activation function to keep the output in the action space of agent. To minimize the loss function of the algorithm, Adam \cite{Adam} is used as the default optimizer for all neural networks to update the samples randomly and uniformly collected for mini-batch 100 with a learning rate of $3e-4$. The actor network and two target critic networks perform delayed soft updates every $d=2$ iterations, where $\tau=0.005$. To balance overestimate and underestimate and to fairly evaluate our algorithm, we use $\beta=0.45$ on all tasks when computing the target critic. The same setting is applied to OurDDPG to fairly compare the estimation of the value function and that of WD3. 

To balance exploration and exploitation, the Gaussian noise of $\epsilon \sim \mathcal{N}(0,0.1)$ is added to the actions when an agent selects actions to interact with the environment, and then the actions with noise are clipped in the action space of the agent. When updating the value function, we add the Gaussian noise of $\epsilon \sim \mathcal{N}(0,0.2)$ to the action selected according to the target actor, which is clipped to $[-0.5, 0.5]$. To eliminate the dependence of the policy network on the initial parameters, we used the pure exploration policy for all environments for the first 25,000 time steps.

Each task runs on 1 million time steps, with evaluations conducted every 5,000 time steps. All algorithms are run and evaluated on ten random seeds. In the evaluation process, there is no exploration noise, and the transitions from the evaluation will not be carried over to the experience replay buffer. Furthermore, all our experiments are reported based on ten random seeds.

\subsection{Experimental results}
We compared our algorithm with the TD3 algorithm and the state-of-the-art policy gradient algorithms PPO, TRPO \cite{TRPO},  and DDPG, which are implemented by the OpenAI Baselines \cite{Baselines}. For the TD3 algorithm, we use the author's implementation. Besides, considering that there are some engineering skills in the DDPG implementation of OpenAI, we implement the DDPG algorithm by ourselves, called OurDDPG, without adding any tricks on the original DDPG that affect the performance of the algorithm.

The learning performance curves are graphed in \cref{performance experiments}. The results demonstrate that WD3 matches or outperforms all other algorithms with a consistent hyper-parameter $\beta$. The better performance can be obtained by fine-tuning $\beta$. To further evaluate the influence of $\beta$ on WD3, we take different $\beta$ and evaluate the performance on the Continuous control tasks. The results are graphed in \cref{various_beta}. There is not a very significant difference in the performance achieved by different $\beta$, which shows that WD3 is robust for hyper-parameter $\beta$. 
\subsection{Estimation error}
To investigate the estimation error, we evaluate the estimation error of OurDDPG, TD3, and WD3 in an Ant environment over 10 random seeds. Every 5,000 time steps we get the average action value of the current agent and the true value estimated by the Monte Carlo method. We take 50 trajectories, and each trajectory contains 1,000 transitions to approximate the true action value. All the experiments are carried out on ten random seeds. We present the results in \cref{Estimatation issue}.

The results show that WD3 does achieve our purpose. In the initial training stage of an agent, the estimation of the value function is stable and then rises gradually, which is close to the real value function. In the learning process of the OurDDPG, the action-value function is greatly overestimated, which makes the performance of the algorithm suffer. The TD3 algorithm overestimates the action value in the initial stage as OurDDPG. With the progress of learning, the algorithm begins to underestimate action value. Note that the value function of TD3 appears a small magnitude of overestimation in some stages of action-value function learning. We argue that the overestimation bias in this curve is caused by the delay effect of neural networks.  When an agent explores a new space, the neural network has been updated for a period, which makes the output of the Q-function network higher than that of the previous network. For the state-action pair that is not explored by the agent, the neural network will have the problem of overestimation, which is caused by the unexpected generalization ability of the neural network, i.e., neural networks generalize learned high action values to unseen ones. The value of unseen state-action pairs is usually less than the explored one. Because the agent does not explore this space, the actions outputted by the policy network are not as good as explored space, thus the action-value network cannot match the returns, and the outputs of the action-value function are overestimated. Once the agent learns for a while in this space of this state, the underestimation problem will occur again, which is consistent with our theoretical analysis. The Q-function learning process of the WD3 algorithm is more stable than DDPG and TD3, without underestimation or huge overestimation. WD3 has a preferable property for value function, which leads to better performance.

\section{Conclusion}
The estimation bias is a crucially important challenge in value-based reinforcement learning. In this paper, we prove that the overestimation and underestimation error exist widely in the deterministic policy gradient algorithm in theory and practice. We discuss the overestimation issue of the combination of action-value function learning and neural network, which is caused by the max operator. And we prove that the underestimation issue does occur both in theory and in practice. In order to reduce the estimation bias of the Q-function, we propose the WD3 algorithm, which makes the updating process of the Q-function more stable and accurate utilizing the weighted average target critic, thus improving the performance. We experimentally demonstrate that WD3 is indeed more stable for updating value functions. Furthermore, experiments show that our algorithm matches or outperforms the state-of-the-art algorithms on continuous control tasks. Estimation bias are also related to exploration in DRL. Overestimation will encourage agents to explore the overestimated area while underestimation makes agents conservative for the underestimated area. An easily overlooked fact is that exploration and estimation bias are entangled with each other in deep reinforcement learning systems, and neither can be ignored. A possible solution is to decouple exploration and estimation bias in deep reinforcement learning. For example, the agent utilizes a policy to explore the environment, and another policy is used to accurately learn the value. The sophisticated nature of the relationship between estimation bias and exploration is left for future research.

\section*{Acknowledgment}
This project was supported by the National Key R\&D Program of China (2017YFC1200601), the National Natural Science Foundation of China (31672325), and the Foundation Strengthening Key Project 021-00, Basic Scientific Research Program of China B022.

\bibliographystyle{unsrtnat}
\bibliography{reference}

\end{document}